\documentclass[10pt,twocolumn,letterpaper]{article}

\usepackage{cvpr}
\usepackage{times}
\usepackage{epsfig}
\usepackage{graphicx}
\usepackage{amsmath}
\usepackage{amssymb}

\usepackage[utf8]{inputenc} % allow utf-8 input
\usepackage[T1]{fontenc}    % use 8-bit T1 fonts
\usepackage{url}            % simple URL typesetting
\usepackage{booktabs}       % professional-quality tables
\usepackage{amsfonts}       % blackboard math symbols
\usepackage{nicefrac}       % compact symbols for 1/2, etc.
\usepackage{microtype}      % microtypography
\usepackage{bbm}

\usepackage{amssymb}
\usepackage{subcaption}
\usepackage{changepage}
\usepackage{epstopdf}
\usepackage[dvipsnames]{xcolor}
\usepackage{colortbl}
\definecolor{LightCyan}{rgb}{0.8,1,1}
\usepackage{mathtools}
\usepackage{amsmath}
\usepackage{setspace}
\usepackage{booktabs}% http://ctan.org/pkg/booktabs
\usepackage{array}
\usepackage{algorithm}
\usepackage{algorithmic}
\usepackage{multirow}
\usepackage[pagebackref=true,breaklinks=true,letterpaper=true,colorlinks,bookmarks=false]{hyperref}

\cvprfinalcopy % *** Uncomment this line for the final submission

\def\cvprPaperID{4179} % *** Enter the CVPR Paper ID here

\newtheorem{ntheorem}{Theorem}

\newtheorem{proposition}[ntheorem]{Proposition}

\newcommand{\argmin}{\operatornamewithlimits{argmin}}

\newcommand{\colvec}[1]{\begin{bmatrix}#1\end{bmatrix}}

\newcommand{\bb}[1]{\mathbf{#1}}

\newcommand{\sumi}[2]{\sum_{{#1}=1}^{#2}}

\def\bal#1\eal{\begin{align*}#1\end{align*}}
\newcommand{\R}{\mathbb{R}}
\newcommand{\T}{\intercal}

\newcommand{\removed}[1]{}

\ifcvprfinal\fi
\begin{document}

%%%%%%%%% TITLE
\title{Object Discovery in Videos as Foreground Motion Clustering}

\author{Christopher Xie$^1$\thanks{Work partially done while an intern at NVIDIA.} \hspace{4px} Yu Xiang$^2$ \hspace{4px} Zaid Harchaoui$^1$ \hspace{4px} Dieter Fox$^{2,1}$\\
$^1$University of Washington \hspace{6px} $^2$NVIDIA\\
\tt\small chrisxie@cs.washington.edu \hspace{2px} \{yux,dieterf\}@nvidia.com \hspace{2px} zaid@uw.edu
}

\maketitle
%\thispagestyle{empty}

%%%%%%%%% ABSTRACT
\begin{abstract}
   We consider the problem of providing dense segmentation masks for object discovery in videos. We formulate the object discovery problem as foreground motion clustering, where the goal is to cluster foreground pixels in videos into different objects. We introduce a novel pixel-trajectory recurrent neural network that learns feature embeddings of foreground pixel trajectories linked across time. By clustering the pixel trajectories using the learned feature embeddings, our method establishes correspondences between foreground object masks across video frames. To demonstrate the effectiveness of our framework for object discovery, we conduct experiments on commonly used datasets for motion segmentation, where we achieve state-of-the-art performance.
  \vspace{-1mm}
\end{abstract}

\section{Introduction}

Discovering objects from videos is an important capability that an intelligent system needs to have. Imagine deploying a robot to a new environment. If the robot can discover and recognize unknown objects in the environment by observing, it would enable the robot to better understand its work space. In the interactive perception setting \cite{bohg2017interactive}, the robot can even interact with the environment to discover objects by touching or pushing objects. To tackle the object discovery problem, we need to answer the question: what defines an object? In this work, we consider an entity that can move or be moved to be an object, which includes various rigid, deformable and articulated objects. We utilize motion and appearance cues to discover objects in videos.

Motion-based video understanding has been studied in computer vision for decades. In low-level vision, different methods have been proposed to find correspondences between pixels across video frames, which is known as optical flow estimation \cite{horn1981determining,barron1994performance}. Both camera motion and object motion can result in optical flow. Since the correspondences are estimated at a pixel level, these methods are not aware of the objects in the scene, in the sense that they do not know which pixels belong to which objects. In high-level vision, object detection and object tracking in videos has been well-studied \cite{babenko2011robust,kalal2012tracking,hare2016struck,zhang2008global,berclaz2011multiple,xiang2015learning}. These methods train models for specific object categories using annotated data. As a result, they are not able to detect nor track unknown objects that have not been seen in the training data. In other words, these methods cannot discover new objects from videos. In contrast, motion segmentation methods \cite{brox2010object,keuper2015motion,bideau2016rubric,pathak2017learning} aim at segmenting moving objects in videos, which can be utilized to discover new objects based on their motion.

In this work, we formulate the object discovery problem as foreground motion clustering, where the goal is to cluster pixels in a video into different objects based on their motion. There are two main challenges in tackling this problem. First, how can foreground objects be differentiated from background? Based on the assumption that moving foreground objects have different motion as the background, we design a novel encoder-decoder network that takes video frames and optical flow as inputs and learns a feature embedding for each pixel, where these feature embeddings are used in the network to classify pixels into foreground or background. Compared to traditional foreground/background segmentation methods \cite{cheng2006flexible,hu2011incremental}, our network automatically learns a powerful feature representation that combines appearance and motion cues from images. 

\begin{figure}[t]
    \centering
    \includegraphics[width=\linewidth]{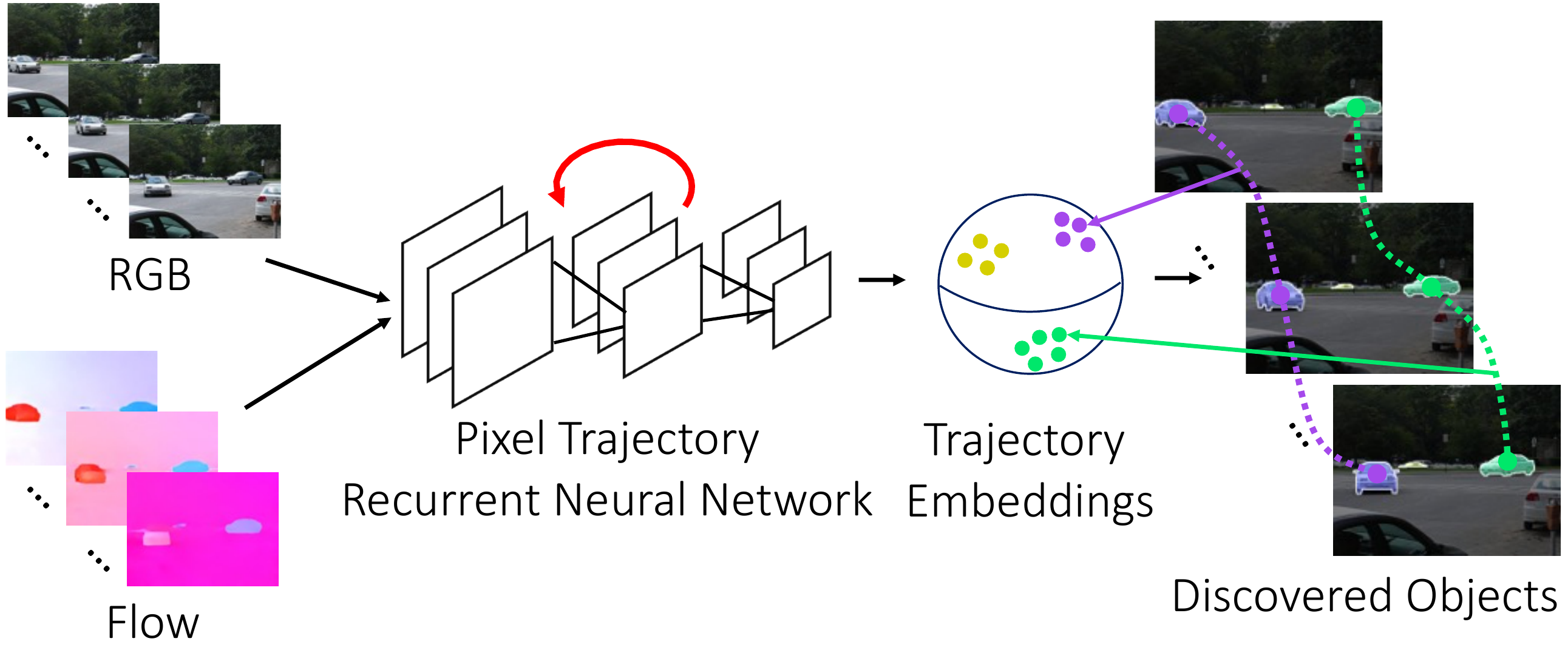}
    \caption{Overview of our framework. RGB images and optical flow are fed into a recurrent neural network, which computes embeddings of pixel trajectories. These embeddings are clustered into different foreground objects. }
    \label{fig:teaser_figure}
    \vspace{-4mm}
\end{figure}

Secondly, how can we consistently segment foreground objects across video frames? We would like to segment individual objects in each video frame and establish correspondences of the same object across video frames. Inspired by \cite{brox2010object} that clusters pixel trajectories across video frames for object segmentation, we propose to learn feature embeddings of pixel trajectories with a novel Recurrent Neural Network (RNN), and then cluster these pixel trajectories with the learned feature embeddings. Since the pixel trajectories are linked in time, our method automatically establishes the object correspondences across video frames by clustering the trajectories. Different from \cite{brox2010object} that employs hand-crafted features to cluster pixel trajectories, our method automatically learns a feature representation of the trajectories, where the RNN controls how to combine pixel features along a trajectory to obtain the trajectory features. Figure \ref{fig:teaser_figure} illustrates our framework for object motion clustering.

Since our problem formulation aims to discover objects based on motion, we conduct experiments on five motion segmentation datasets to evaluate our method: Flying Things 3D \cite{MIFDB16}, DAVIS \cite{Perazzi2016, Pont-Tuset_arXiv_2017}, Freiburg-Berkeley motion segmentation \cite{ochs2014segmentation}, ComplexBackground \cite{narayana2013coherent} and CamouflagedAnimal \cite{bideau2016s}. We show that our method is able to segment potentially unseen foreground objects in the test videos and consistently across video frames. Comparison with the state-of-the-art motion segmentation methods demonstrates the effectiveness of our learned trajectory embeddings for object discovery. In summary, our work has the following key contributions:
\begin{itemize}
    \item We introduce a novel encoder-decoder network to learn feature embeddings of pixels in videos that combines appearance and motion cues.
    \item We introduce a novel recurrent neural network to learn feature embeddings of pixel trajectories in videos.
    \item We use foreground masks as an attention mechanism to focus on clustering of relevant pixel trajectories for object discovery.
    \item We achieve state-of-the-art performance on commonly used motion segmentation datasets.
\end{itemize}

This paper is organized as follows. After discussing related work, we introduce our foreground motion clustering method designed for object discovery, followed by experimental results and a conclusion.
\section{Related Work}

\noindent \textbf{Video Foreground Segmentation.} Video foreground segmentation is the task of classifying every pixel in a video as foreground or background. This has been well-studied in the context of video object segmentation \cite{bideau2016s, papazoglou2013fast, tokmakov2017mpnet, tokmakov2017mpnet, fusionseg}, especially with the introduction of unsupervised challenge of the DAVIS dataset \cite{Perazzi2016}. \cite{bideau2016s} uses a probabilistic model that acts upon optical flow to estimate moving objects. \cite{papazoglou2013fast} predicts video foreground by iteratively refining motion boundaries while encouraging spatio-temporal smoothness. \cite{tokmakov2017mpnet, tokmakov2017learning, fusionseg} adopt a learning-based approach and train Convolutional Neural Networks (CNN) that utilize RGB and optical flow as inputs to produce foreground segmentations. Our approach builds on these ideas and uses the foreground segmentation as an attention mechanism for pixel trajectory clustering.

\vspace{2mm}
\noindent \textbf{Instance Segmentation.} Instance segmentation algorithms segment individual object instances in images. Many instance segmentation approaches have adopted the general idea of combining segmentation with object proposals \cite{he2017mask, SharpMask}. While these approaches only work for objects that have been seen in a training set, we make no such assumption as our intent is to discover objects. Recently, a few works have investigated the instance segmentation problem as a pixel-wise labeling problem by learning pixel embeddings \cite{de2017semantic, Novotny_2018_ECCV, kong2018recurrent, fathi2017semantic}. \cite{Novotny_2018_ECCV} predicts pixel-wise features using translation-variant semi-convolutional operators. \cite{fathi2017semantic} learns pixel embeddings with seediness scores that are used to compose instance masks. \cite{de2017semantic} designs a  contrastive loss and \cite{kong2018recurrent} unrolls mean shift clustering as a neural network to learn pixel embeddings. We leverage these ideas to design our approach of learning embeddings of pixel trajectories.

% While the output of instance segmentation algorithms is a dense segmentation mask of an unknown number of instances, which is similar to our work.

\vspace{2mm}
\noindent \textbf{Motion Segmentation.} Pixel trajectories for motion analysis were first introduced by \cite{sundaram2010dense}. \cite{brox2010object} used them in a spectral clustering method to produce motion segments. \cite{ochs2014segmentation} provided a variational minimization to produce pixel-wise motion segmentations from trajectories. Other works that build off this idea include formulating trajectory clustering as a multi-cut problem \cite{Keuper2017HigherOrderMC, keuper2015motion, keuper2018motion} or as a density peaks clustering \cite{wang2017super}, and detecting discontinuities in the trajectory spectral embedding \cite{fragkiadaki2012video}. More recent approaches include using occlusion relations to produce layered segmentations \cite{taylor2015causal}, combining piecewise rigid motions with pre-trained CNNs to merge the rigid motions into objects \cite{bideau2018best}, and jointly estimating scene flow and motion segmentations \cite{2018_RAL_ssdb}. We use pixel trajectories in a recurrent neural network to learn trajectory embeddings for motion clustering. 

%Differing from these past works, we take a novel learning-based approach to the problem and provide exceptional results.

%Due to the inconsistency and ambiguity of motion segmentation dataset labels, \cite{bideau2016rubric} proposes a detailed formulation in order to structure the problem. They then provided corrected groundtruth labels for a few datasets, which we use in Seciton \ref{sec:experiments}. \textcolor{red}{While our problem formulation is slightly different from the specifications in \cite{bideau2016rubric} (according to their rubrik, our formulation is more like segmentation and tracking of objects), we can still evaluate our methods on their ruleset.}

\section{Method}

\begin{figure*}[t]
\begin{center}
\includegraphics[width=\linewidth]{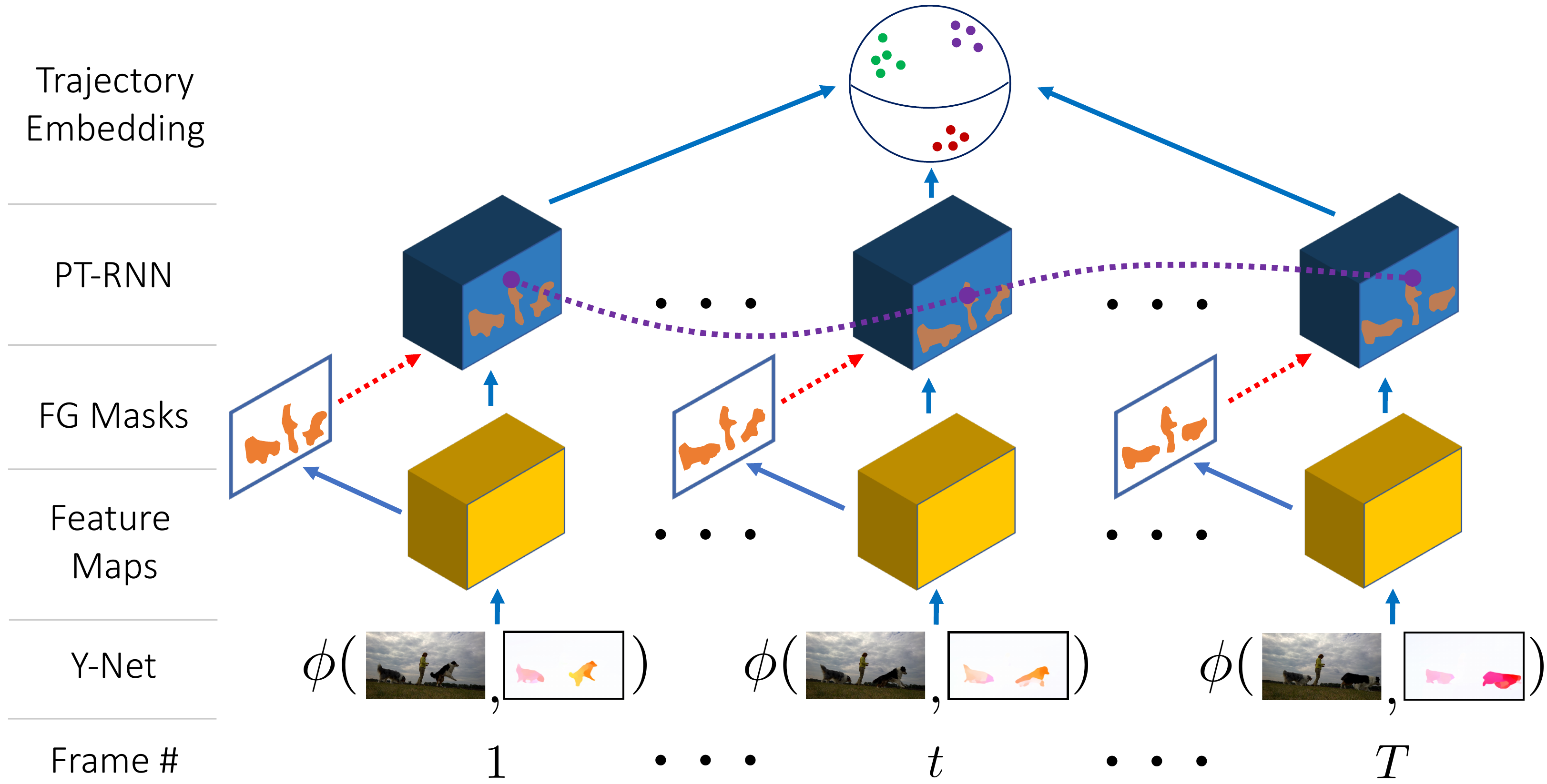}
\caption{Overall architecture. First, feature maps of each frame are extracted from the Y-Net. Next, foreground masks are computed, shown in \textcolor{orange}{orange}. The PT-RNN uses these foreground masks to compute trajectory embeddings (example foreground trajectory from frame $1$ to $T$ shown in \textcolor{purple}{purple}), which are normalized to produce unit vectors. Backpropagation passes through the \textcolor{blue}{blue} solid arrows, but not through the \textcolor{red}{red} dashed arrows.}
\label{fig:pt_rnn_architecture}
\vspace{-6mm}
\end{center}
\end{figure*}

Our approach takes video frames and optical flow between pairs of frames as inputs, which are fed through an encoder-decoder network, resulting in pixel-wise features. These features are used to predict foreground masks of moving objects. In addition, a recurrent neural network is designed to learn feature embeddings of pixel trajectories inside the foreground masks. Lastly, the trajectory embeddings are clustered into different objects, giving a consistent segmentation mask for each discovered object. The network architecture is visualized in Figure \ref{fig:pt_rnn_architecture}. %We describe the details of our method in this section.

\subsection{Encoder-Decoder: Y-Net}

%In this work, we introduce a variation of the U-Net architecture that learns a novel mid-level fusion of low-resolution features. We call this network Y-Net. 

\begin{figure}
\begin{center}
\begin{subfigure}[t]{0.49\linewidth}
% \includegraphics[width=\linewidth]{Images/U-Net.pdf}

% U-Net pdf is smaller, so make it align vertically with Y-Net pdf
\setbox1=\hbox{\includegraphics[width=\linewidth]{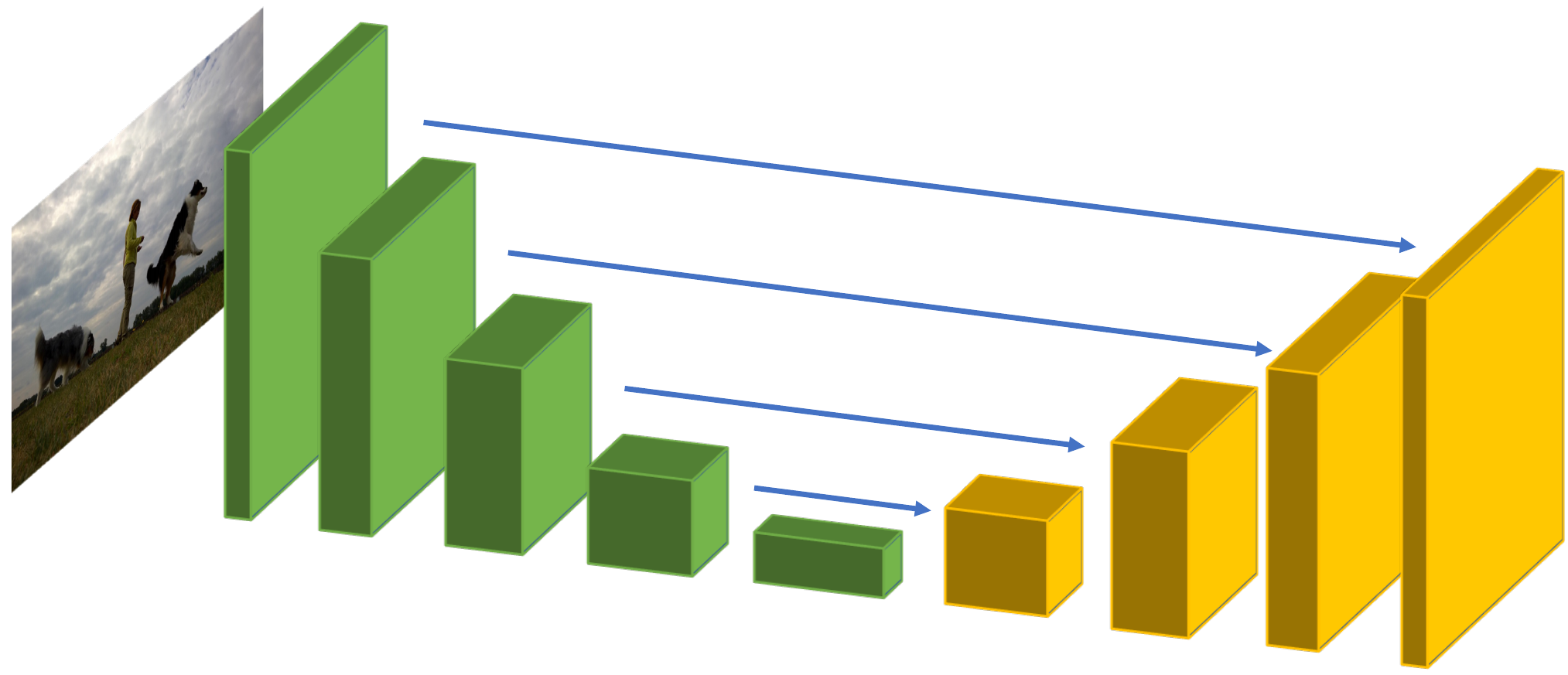}}% The smaller image
\setbox2=\hbox{\includegraphics[width=\linewidth]{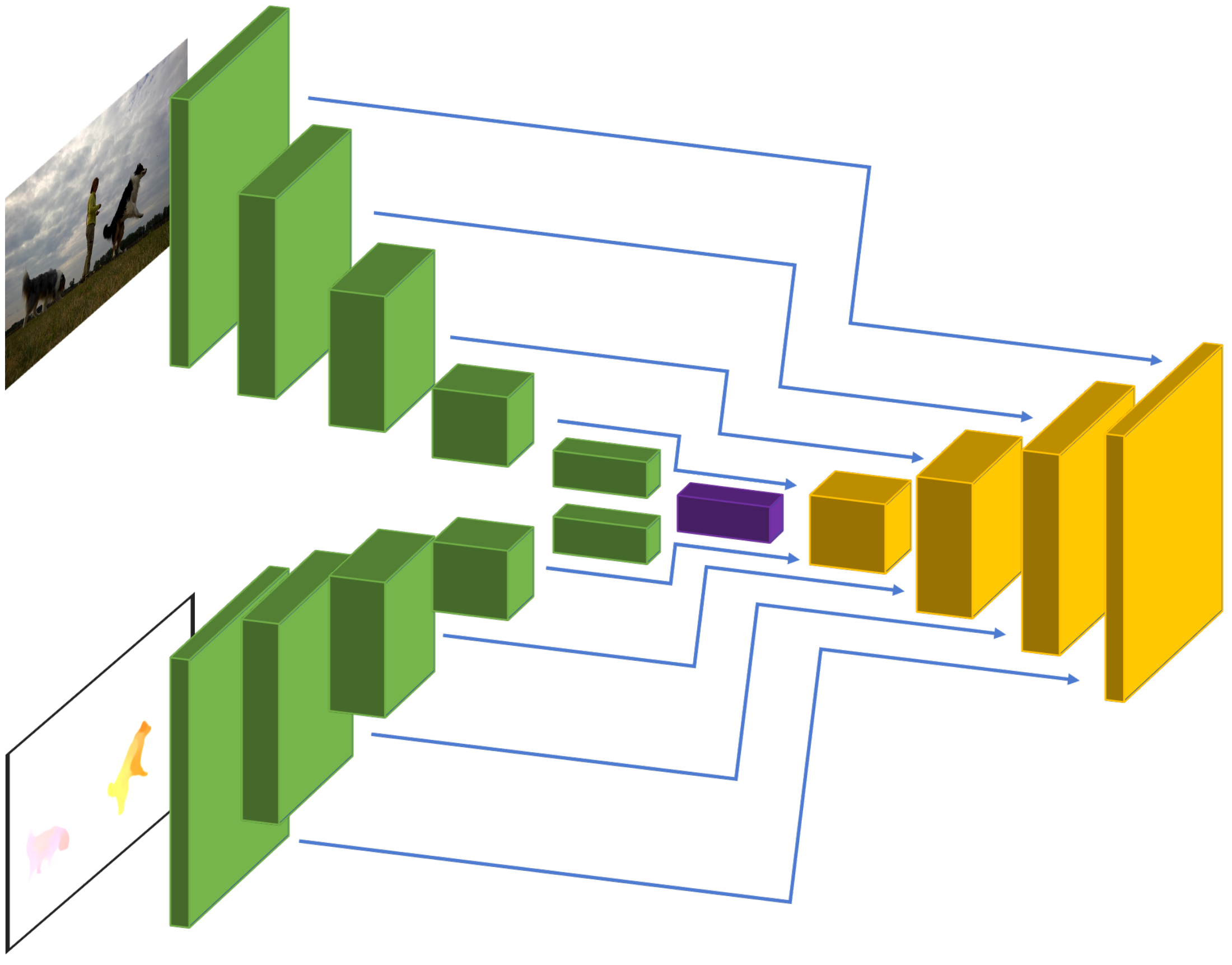}}% The larger image
\raisebox{0.5\ht2-0.5\ht1}{\includegraphics[width=\linewidth]{Images/U-Net.pdf}}

\caption{U-Net architecture}
\label{fig:unet_arch}
\end{subfigure}
\hfill
\begin{subfigure}[t]{0.49\linewidth}
\includegraphics[width=\linewidth]{Images/Y-Net.pdf}
\caption{Y-Net architecture}
\label{fig:ynet_arch}
\end{subfigure}

\caption{We show U-Net \cite{ronneberger2015u} and our proposed Y-Net to visually demonstrate the difference. Y-Net has two encoding branches (shown in \textcolor{green}{green}) for each input modality, which is fused (shown in \textcolor{purple}{purple}) and passed to the decoder (shown in \textcolor{yellow}{yellow}). Skip connections are visualized as \textcolor{blue}{blue} arrows.}
\vspace{-6mm}
\label{fig:unet_ynet_architecture_comparison}
\end{center}
\end{figure}

Let $I_t \in \R^{H \times W \times 3}, F_t \in \R^{H \times W \times 2}$ be an RGB image and forward optical flow image at time $t$, respectively. Our network receives these images from a video as inputs and feeds them into an encoder-decoder network separately at each time step, where the encoder-decoder network extracts dense features for each video frame. Our encoder-decoder network is an extension of the U-Net architecture \cite{ronneberger2015u} (Figure \ref{fig:unet_arch}) to two different input types, i.e., RGB images and optical flow images, by adding an extra input branch. We denote this mid-level fusion of low-resolution features as Y-Net. We illustrate the Y-Net architecture in Figure \ref{fig:ynet_arch}. 

In detail, our network has two parallel encoder branches for the RGB and optical flow inputs. Each encoder branch consists of four blocks of two $3\times 3$ convolutions (each of which is succeeded by a GroupNorm layer \cite{wu2018group} and ReLU activation) followed by a $2\times 2$ max pooling layer. The encodings of the RGB and optical flow branches are then concatenated and input to a decoder network, which consists of a similar architecture to \cite{ronneberger2015u} with skip connections from both encoder branches to the decoder.

We argue that this mid-level fusion performs better than early fusion and late fusion (using completely separate branches for RGB and optical flow, similar to two-stream networks \cite{simonyan2014two, feichtenhofer2016convolutional, tokmakov2017learning}) of encoder-decoder networks while utilizing less parameters, and show this empirically in Section \ref{subsec:ablation_studies}. The output of Y-Net, $\phi(I_t, F_t) \in \R^{H \times W \times C}$, is a pixel-dense feature representation of the scene. We will refer to this as pixel embeddings of the video.

\subsection{Foreground Prediction}
 
The Y-Net extracts a dense feature map for each video frame that combines appearance and motion information of the objects. Using these features, our network predicts a foreground mask for each video frame by simply applying another convolution on top of the Y-Net outputs to compute foreground logits. These logits are passed through a sigmoid layer and thresholded at 0.5. For the rest of the paper, we will denote $m_t$ to be the binary foreground mask at time $t$.

The foreground masks are used as an attention mechanism to focus on the clustering of the trajectory embeddings. This results in more stable performance, as seen in Section \ref{subsec:ablation_studies}. Note that while we focus on moving objects in our work, the foreground can be specified depending on the problem. For example, if we specify that certain objects such as cars should be foreground, then we would learn a network that learns to discover and segment car instances in videos.

\subsection{Trajectory Embeddings}

In order to consistently discover and segment objects across video frames, we propose to learn deep representations of {\it foreground pixel trajectories} of the video. Specifically, we consider dense pixel trajectories throughout the videos, where trajectories are defined as in \cite{sundaram2010dense, brox2010object}. Given the outputs of Y-Net, we compute the trajectory embedding as a weighted sum of the pixel embeddings along the trajectory.

\subsubsection{Linking Foreground Trajectories}

We first describe the method to calculate pixel trajectories according to \cite{sundaram2010dense}. Denote $F_{t-1} \in \R^{H \times W \times 2}$ to be the forward optical flow field at time $t-1$ and $\hat{F}_t \in \R^{H \times W \times 2}$ to be the backward optical flow field at time $t$. As defined in \cite{sundaram2010dense}, we say the optical flow for two pixels $(i,j)$ at time $t-1$ and $(i',j')$ at time $t$ is consistent if
\begin{equation} \label{eq:flow_consistency}
    \left\| F_{t-1}^{i,j} + \hat{F}_t^{i',j'} \right\|^2 \leq 0.01 \left( \left\| F_{t-1}^{i,j} \right\|^2 + \left\| \hat{F}_t^{i',j'} \right\|^2 \right) + 0.5,
\end{equation}
where $F_{t-1}^{i,j}$ denotes the $i,j$-th element of $F_{t-1}$. Essentially, this condition requires that the backward flow points in the inverse direction of the forward flow, up to a tolerance interval that is linear in the magnitude of the flow. Pixels $(i,j)$ and $(i', j')$ are linked in a pixel trajectory if Eq. (\ref{eq:flow_consistency}) holds. 

\begin{figure}
\begin{center}
\includegraphics[width=\linewidth]{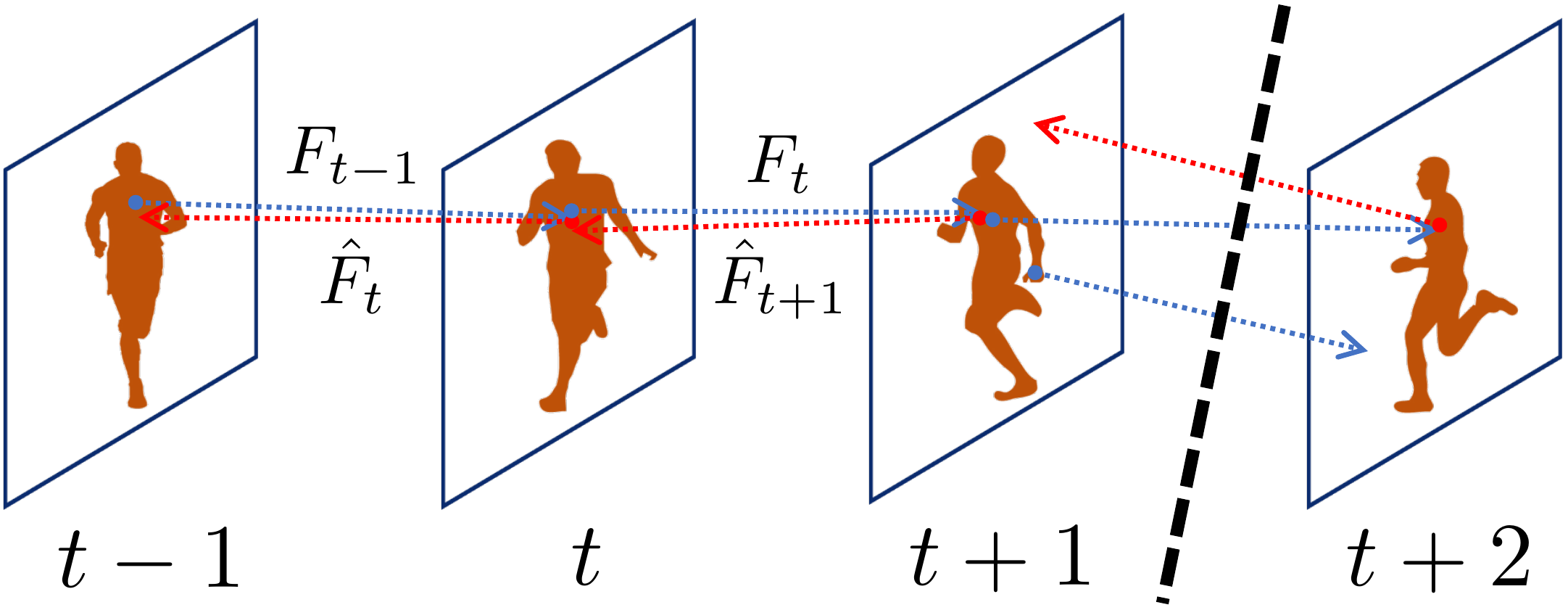}
\caption{We illustrate pixel linking in foreground pixel trajectories. The foreground mask is shown in \textcolor{orange}{orange}, forward flow is denoted by the \textcolor{blue}{blue} dashed arrow, and backward flow is denoted by the \textcolor{red}{red} dashed arrow. The figure shows a trajectory that links pixels in frames $t-1, t, t+1$. Two failure cases that can cause a trajectory to end are shown between frames $t+1$ and $t+2$: 1) Eq. (\ref{eq:flow_consistency}) is not satisfied, and 2) one of the pixels is not classified as foreground.}
\label{fig:fg_trajectory}
\vspace{-8mm}
\end{center}
\end{figure}

To define {\it foreground pixel trajectories}, we augment the above definition and say pixels $(i,j)$ and $(i',j')$ are linked if Eq. (\ref{eq:flow_consistency}) holds and both pixels are classified as foreground. Using this, we define a foreground-consistent warping function $g : \R^{H \times W} \rightarrow \R^{H \times W}$ that warps a set of pixels $\bb{v} \in \R^{H \times W}$ forward in time along their foreground trajectories:
\vspace{-1mm}
\begin{displaymath}
g(\bb{v})^{i',j'} = \left \{
    \begin{array}{lr}
        \bb{v}^{i,j} & \textrm{if $(i,j)$, $(i',j')$ linked}  \\
        0 & \textrm{otherwise.} 
    \end{array}
    \right.
    \vspace{-1mm}
\end{displaymath}
This can be achieved by warping $\bb{v}$ with $\hat{F}_t$ with bilinear interpolation and multiplying by a binary consistency mask. This mask can be obtained by warping the foreground mask $m_{t-1}$ with $\hat{F}_t$ using Eq. (\ref{eq:flow_consistency}) and intersecting it with $m_t$, resulting in a mask that is 1 if $(i',j')$ is linked to a foreground pixel at time $t-1$. Figure \ref{fig:fg_trajectory} demonstrates the linking of pixels in a foreground pixel trajectory. 

\subsubsection{Pixel Trajectory RNN}
\label{subsubsec:pixel_trajectory_rnn}

After linking foreground pixels into trajectories, we describe our proposed Recurrent Neural Network (RNN) to learn feature embedings of these trajectories. Denote $\{(i_t, j_t)\}_{t=1}^L$ to be the pixel locations of a foreground trajectory, $\{\mathbf{x}_t^{i_t, j_t} \in \R^C\}_{t=1}^L$ to be the pixel embeddings of the foreground trajectory (Y-Net outputs, i.e. $\mathbf{x}_t = \phi(I_t, F_t)$), and $L$ as the length of the trajectory. We define the foreground trajectory embeddings to be a weighted sum of the pixel embeddings along the foreground trajectory. Specifically, we have
\begin{equation} \label{eq:traj_embedding}
    \psi\left(\{\mathbf{x}_t^{i_t, j_t}\}_{t=1}^L\right) = \frac{\sumi{t}{L} \mathbf{w}_{t}^{i_t, j_t} \odot \mathbf{x}_{t}^{i_t, j_t}} {\sumi{t}{L} \mathbf{w}_t^{i_t, j_t}} \; ,
\end{equation}
where $\odot$ denotes element-wise multiplication, the division sign denotes element-wise division, and $\mathbf{w}_t^{i_t, j_t} \in [0,1]^{C}$.

To compute the trajectory embeddings, we encode $\psi(\cdot)$ as a novel RNN architecture which we denote Pixel Trajectory RNN (PT-RNN). In its hidden state, PT-RNN stores
\begin{equation} \label{eq:pt_rnn_hidden_state}
    \left\{\mathbf{h}_t^{i_t, j_t} := \sumi{\tau}{t} \mathbf{w}_{\tau}^{i_{\tau}, j_{\tau}} \odot \mathbf{x}_{\tau}^{i_{\tau}, j_{\tau}} , \mathbf{W}_t^{i_t, j_t} := \sumi{\tau}{t} \mathbf{w}_{\tau}^{i_{\tau}, j_{\tau}} \right\},
\end{equation}
which allows it to keep track of the running sum and total weight throughout the foreground trajectory. While Eq.~\eqref{eq:pt_rnn_hidden_state} describes the hidden state at each pixel location and time step, we can efficiently implement the PT-RNN for all pixels by doing the following: at time step $t$, PT-RNN first applies the foreground consistent warping function to compute $\widetilde{\mathbf{h}}_{t-1} := g\left(\mathbf{h}_{t-1} \right), \widetilde{\mathbf{W}}_{t-1} = g\left( \mathbf{W}_{t-1} \right)$. Next, we compute $\bb{w}_t$. We design three variants of PT-RNN to compute $\bb{w}_t$, named \textit{standard} (based on simple RNNs), \textit{conv} (based on convRNNs), and \textit{convGRU} (based on \cite{ballas2015delving}). For example, our \textit{conv} architecture is described by
\begin{equation}
\begin{split}
\bb{c}_t &= \textrm{ReLU}\left(W_c * \colvec{\frac{\widetilde{\bb{h}}_{t-1}}{\widetilde{\bb{W}}_{t-1}} & \bb{x}_t}\right) \\
\bb{w}_t &= \sigma\left(W_w * \bb{c}_t \right),
\end{split}
\end{equation}
where $*$ denotes convolution, and $W_c, W_w$ are $3\times 3$ convolution kernels. After $\mathbf{w}_t$ is computed by the PT-RNN, we update the hidden state with:
\begin{equation}
    \begin{split}
        \mathbf{h}_t &= \widetilde{\mathbf{h}}_{t-1} + \mathbf{w}_t \odot \mathbf{x}_t \\
        \mathbf{W}_t &= \widetilde{\mathbf{W}}_{t-1} + \mathbf{w}_t.
    \end{split}
\end{equation}
All model variants are described in detail in the supplement. Essentially, {\it standard} treats each set of linked pixels as a simple RNN, {\it conv} includes information from neighboring pixels, and {\it convGRU} allows the network to capture longer term dependencies by utilizing an explicit memory state.

When a trajectory is finished, i.e., pixel $(i,j)$ does not link to any pixel in the next frame, PT-RNN outputs 
$\mathbf{h}_t^{i, j} / \mathbf{W}_t^{i, j}$,
%$\frac{\mathbf{h}_t^{i, j}}{\mathbf{W}_t^{i, j}}$, 
which is equivalent to Eq.~\eqref{eq:traj_embedding}. This results in a $C$-dimensional embedding for every foreground pixel trajectory, regardless of its length, when it starts, or when it ends. Note that these trajectory embeddings are pixel-dense, removing the need for a variational minimization step \cite{ochs2014segmentation}. The embeddings are normalized so that they lie on the unit sphere. 

A benefit to labeling the trajectories is that we are enforcing consistency in time, since consistent forward and backward optical flow usually means that the pixels are truely linked \cite{sundaram2010dense}. However, issues can arise around the motion and object boundaries, which can lead to trajectories erroneously drifting and representing motion of two different objects or an object and background \cite{sundaram2010dense}. In this case, the foreground masks are beneficial and able to sever the trajectory before it drifts. We also note the similarity to the DA-RNN architecture \cite{xiang2017rnn} that uses data association in a RNN for semantic labeling.

%except that our data association comes from consistent optical flow instead of 3D models and our weighted sum is not an exponentially weighted sum. 

\vspace{-2mm}
\subsubsection{Spatial Coordinate Module}

The foreground trajectory embeddings incorporate information from the RGB and optical flow images. However, they do not encode information about the location of the trajectory in the image. Thus, we introduce a spatial coordinate module which computes location information for each foreground trajectory. Specifically, we compute a 4-dimensional vector consisting of the average $x,y$ pixel location and displacement for each trajectory and pass it through two fully connected (FC) layers to inflate it to a $C$-dimensional vector, which we add to the output of $\psi(\cdot)$ (before the normalization of the foreground trajectory embeddings).

\subsection{Loss Function}

To train our proposed network, we use a loss function that is comprised of three terms
\begin{equation*} \label{eq:loss}
    \mathcal{L} = \lambda_{\textrm{fg}} \ell_{\textrm{fg}} + \lambda_{\textrm{intra}} \ell_{\textrm{intra}} + \lambda_{\textrm{inter}} \ell_{\textrm{inter}} \; ,
\end{equation*}
where we set $\lambda_{\textrm{fg}} = \lambda_{\textrm{intra}} = \lambda_{\textrm{inter}} = 1$ in our experiments. $\ell_{\textrm{fg}}$ is a pixel-wise binary cross-entropy loss that is commonly used in foreground prediction. We apply this on the predicted foreground logits. $\ell_{\textrm{intra}}$ and $\ell_{\textrm{inter}}$ operate on the foreground trajectory embeddings. Inspired by \cite{de2017semantic}, its goal is to encourage trajectory embeddings of the same object to be close while pushing trajectories that are different objects apart. For simplicity of notation, let us overload notation and define $\left\{\mathbf{x}_i^k\right\}, k=1,\ldots,K,\ i=1,\ldots,N_k$ to be a list of trajectory embeddings of dimension $C$ where $k$ indexes the object and $i$ indexes the embedding. Since all the feature embeddings are normalized to have unit length, we use the cosine distance function
$d(\mathbf{x},\mathbf{y}) = \frac12 \left(1 - \mathbf{x}^\T \mathbf{y}\right)$
to measure the distance between two feature embeddings $\mathbf{x}$ and $\mathbf{y}$.
%
%\begin{equation}
%    d(\mathbf{x},\mathbf{y}) = \frac12 \left(1 - \mathbf{x}^\T \mathbf{y}\right) \in [0,1].
%\end{equation}

%Note that $d$ is not a metric.

\begin{proposition} \label{prop:spherical_mean}
    Let $\{\mathbf{y}_i\}_{i=1}^N$ be a set of unit vectors such that $\sumi{i}{n} \bb{y}_i \neq 0$. Define the \emph{spherical mean} of this set of unit vectors to be the unit vector that minimizes the cosine distance
    \begin{equation}
        \mathbf{\mu} := \argmin_{\|\mathbf{w}\|_2 = 1} \frac1n \sumi{i}{n} d\left(\mathbf{w},\mathbf{y}_i\right) \; 
    \end{equation}
    Then $\mathbf{\mu} = \frac{\sumi{1}{n} \mathbf{y}_i}{\left \|\sumi{1}{n} \mathbf{y}_i \right \|}$. For the proof, see the supplement.
\end{proposition}

The goal of the intra-object loss $\ell_{\textrm{intra}}$ is to encourage these learned trajectory embeddings of an object to be close to their spherical mean. This results in 
\begin{equation*} \label{eq:lambda_intra}
    \ell_{\textrm{intra}} = \frac1K \sum_{k=1}^K \sumi{i}{N_k} \frac{\mathbbm{1}\left\{d(\mu_k, \mathbf{x}_i^k) - \alpha \geq 0\right\}\ d^2(\mu_k, \mathbf{x}_i^k)}{\sumi{i}{N_k} \mathbbm{1}\left\{d(\mu_k, \mathbf{x}_i^k) - \alpha \geq 0\right\}} \; ,
\end{equation*}
where $\mu_k$ is the spherical mean of trajectories $\left\{\mathbf{x}_i^k\right\}_{i=1}^{N_k}$ for object $k$, and $\mathbbm{1}$ denotes the indicator function. Note that $\mu_k$ is a function of the embeddings. The indicator function acts as a hard negative mining that focuses the loss on embeddings that are further than margin $\alpha$ from the spherical mean. In practice, we do not let the denominator get too small as it could result in unstable gradients, so we allow it to reach a minimum of 50. 

Lastly, the inter-object loss $\ell_{\textrm{inter}}$ is designed to push trajectories of different objects apart. We desire the clusters to be pushed apart by some margin $\delta$, giving
\begin{equation*} \label{eq:lambda_inter}
    \ell_{\textrm{inter}} = \frac{2}{K(K-1)} \sum_{k < k'} \left[\delta - d(\mu_k, \mu_{k'}) \right]_+^2 \; ,
\end{equation*}
where $[x]_+ = \max(x,0)$. This loss function encourages the spherical means of different objects to be at least $\delta$ away from each other. Since our embeddings lie on the unit sphere and our distance function measures cosine distance, $\delta$ does not need to depend on the feature dimension $C$. In our experiments, we set $\delta=0.5$ which encourages the clusters to be at least 90 degrees apart.

\subsection{Trajectory Clustering}

At inference time, we cluster the foreground trajectory embeddings with the von Mises-Fisher mean shift (vMF-MS) algorithm \cite{kobayashi2010mises}. This gives us the clusters as well as the number of clusters, which is the estimated number of objects in a video. vMF-MS finds the modes of the kernel density estimate using the von Mises-Fisher distribution. The density can be described as $p(\mathbf{y}; \mathbf{m}, \kappa) = C(\kappa) \exp\left( \kappa \mathbf{m}^\T \mathbf{y} \right)$ for unit vector $\mathbf{y}$ where $\kappa$ is a scalar parameter, $\|\mathbf{m}\|_2 = 1$, and $C(\kappa)$ is a normalization constant. $\kappa$ should be set to reflect the choice of $\alpha$. If the training loss is perfect and $d(\mu_k, \mathbf{x}_i^k) < \alpha, \forall i=1, \ldots, N_k$, then all of the $\mathbf{x}_i^k$ lie within a ball with angular radius $\cos^{-1}(1-2\alpha)$ of $\mu_k$. In our experiments, we set $\alpha = 0.02$, giving $\cos^{-1}(1-2\alpha) \approx 16$ degrees. Thus, we set $\kappa = 10$, resulting in almost 50\% of the density being concentrated in a ball with radius 16 degrees around $\mathbf{m}$ (by eyeing Figure 2.12 of \cite{straub2017phdthesis}). 

\removed{Note that because the trajectory embeddings have no notion of length, start, or end, we compare these trajectories regardless if they overlap in time or not, differing from \cite{brox2010object}. This formulation allows for potential object re-identification, where objects leave and re-enter the scene.}

Running the full vMF-MS clustering is inefficient due to our trajectory representation being pixel-dense. Instead, we run the algorithm on a few randomly chosen seeds that are far apart in cosine distance. If the network learns to correctly predict clustered trajectory embeddings, then this random initialization should provide little variance in the results. Furthermore, we use a PyTorch-GPU implementation of the vMF-MS clustering for efficiency. 
\section{Experiments}
\label{sec:experiments}

\begin{table}[t]
\centering
\small
\begin{tabular}{c|c|c|c}
     & FT3D & DAVIS & FBMS \\ \hline
    Y-Net & \underline{0.905} & \underline{0.701} & \underline{0.631} \\
    Early Fusion & 0.883 & 0.636 & 0.568 \\
    Late Fusion & 0.897 & 0.631 & 0.570 \\ \hline
\end{tabular}
\caption{Fusion ablation. Performance is measured in IoU.}
\label{table:fusion_ablation}
\vspace{-4mm}
\end{table}

\noindent \textbf{Datasets.} We evaluate our method on video foreground segmentation and multi-object motion segmentation on five datasets: Flying Things 3d (FT3D) \cite{MIFDB16}, DAVIS2016 \cite{Perazzi2016}, Freibug-Berkeley motion segmentation \cite{ochs2014segmentation}, Complex Background \cite{narayana2013coherent}, and Camouflaged Animal \cite{bideau2016s}. For FT3D, we combine object segmentation masks with foreground labels provided by \cite{tokmakov2017mpnet} to produce motion segmentation masks. For DAVIS2016, we use the $\mathcal{J}$-measure and $\mathcal{F}$-measure for evaluation. For FBMS, Complex Background, and Camouflaged Animal, we use precision, recall, and F-score, and $\Delta$Obj metrics for evaluation as defined in \cite{ochs2014segmentation, bideau2018best}. Full details of each dataset can be found in the supplement.

It is well-understood that the original FBMS labels are ambiguous \cite{bideau2016rubric}. Some labels exhibit multiple segmentations for one aggregate motion, or segment the (static) background into multiple regions. Thus, \cite{bideau2016rubric} provides corrected labels which we use for evaluation. 

\begin{table}[t]
\resizebox{\linewidth}{!}{\begin{tabular}{c|cccc|ccc}
\hline
& \multicolumn{4}{c|}{Multi-object} & \multicolumn{3}{c}{Foreground} \\ \hline
 & \textcolor{orange}{P} & \textcolor{cyan}{R} & \textcolor{purple}{F} & $\Delta$Obj & \textcolor{orange}{P} & \textcolor{cyan}{R} & \textcolor{purple}{F} \\ \hline \hline
{\it conv} PT-RNN & 75.9 & \underline{66.6} & \underline{67.3} & 4.9 & 90.3 & 87.6 & 87.7 \\
{\it standard} PT-RNN & 72.2 & \underline{66.6} & 66.0 & 4.27 & 88.1 & \underline{89.3} & 87.5 \\
{\it convGRU} PT-RNN & 73.6 & 63.8 & 64.8 & 4.07 & 89.6 & 85.8 & 86.3 \\
per-frame embedding & \underline{79.9} & 56.7 & 59.7 & 11.2 & \underline{92.1} & 85.4 & 87.4 \\
no FG mask & 63.5 & 60.3 & 59.6 & \underline{1.97} & 82.5 & 85.7 & 82.1\\
no SCM & 70.4 & 65.5 & 63.2 & 3.70 & 89.3 & 89.1 & \underline{88.1} \\ \hline \hline
no pre-FT3D & 70.2 & 63.6 & 63.1 & 3.66 & 87.6 & 88.2 & 86.3 \\
no DAVIS-m & 66.9 & 63.6 & 62.1 & 2.07 & 87.1 & 86.9 & 85.2 \\ \hline
\end{tabular}}
\caption{Architecture and Dataset ablation on FBMS testset.}
\label{table:architecture_and_dataset_ablation}
\vspace{-4mm}
\end{table}

\vspace{2mm}
\noindent \textbf{Implementation Details.} We train our networks using stochastic gradient descent with a fixed learning rate of 1e-2. We use backpropagation through time with sequences of length 5 to train the PT-RNN. Each image is resized to $224 \times 400$ before processing. During training (except for FT3D), we perform data augmentation, which includes translation, rotation, cropping, horizontal flipping, and color warping. We set $C=32, \alpha=0.02, \delta=0.5, \kappa=10$. We extract optical flow via \cite{IMKDB17}. 

Labels for each foreground trajectory are given by the frame-level label of the last pixel in the trajectory. Due to sparse labeling in the FBMS training dataset, we warp the labels using Eq. (\ref{eq:flow_consistency}) so that each frame has labels. Lastly, due to the small size of FBMS (29 videos for training), we leverage the DAVIS2017 dataset \cite{Pont-Tuset_arXiv_2017} and hand select 42 videos from the 90 videos that roughly satisfy the rubric of \cite{bideau2016rubric} to augment the FBMS training set. We denote this as DAVIS-m. The exact videos in DAVIS-m can be found in the supplement. 

When evaluating the full model on long videos, we suffer from GPU memory constraints. Thus, we devise a sliding window scheme to handle this. First, we cluster all foreground trajectories within a window. We match the clusters of this window with the clusters of the previous window using the Hungarian algorithm. We use distance between cluster centers as our matching cost, and further require that matched clusters must have $d(\mu_k, \mu_{k'}) < 0.2$. When a cluster is not matched to any of the previous clusters, we declare it a new object. We use a 5-frame window and adopt this scheme for the FBMS and Camouflaged Animal datasets.

In Section \ref{subsec:SOTA_comparison}, we use the {\it conv} PT-RNN variant of Figure \ref{fig:pt_rnn_architecture}, trained for 150k iterations on FT3D, then fine-tuned on FBMS+DAVIS-m for 100k iterations. 

Our implementation is in PyTorch, and all experiments run on a single NVIDIA TitanXP GPU. Given optical flow, our algorithm runs at approximately 15 FPS. Note that we do not use a CRF post-processing step for motion segmentation. 

\begin{table*}[t]
\resizebox{\linewidth}{!}{\begin{tabular}{|cc|cccccc|c|ccc|c|}
\hline
 & & \multicolumn{7}{c|}{Video Foreground Segmentation} & \multicolumn{4}{c|}{Multi-object Motion Segmentation} \\ \hline
 &  & PCM \cite{bideau2016s} & FST \cite{papazoglou2013fast} & NLC \cite{faktor2014video} & MPNet \cite{tokmakov2017mpnet} & LVO \cite{tokmakov2017learning} & CCG \cite{bideau2018best} & Ours & CVOS \cite{taylor2015causal} & CUT \cite{keuper2015motion} & CCG \cite{bideau2018best} & Ours \\ \hline \hline
\multirow{4}{*}{\rotatebox[origin=c]{90}{FBMS}} & \textcolor{orange}{P} & 79.9 & 83.9 & 86.2 & 87.3 & \textcolor{red}{92.4} & 85.5 & \textcolor{blue}{90.3} & 72.7 & \textcolor{blue}{74.6} & 74.2 & \textcolor{red}{75.9} \\
  & \textcolor{cyan}{R} & 80.8 & 80.0 & 76.3 & 72.2 & \textcolor{blue}{85.1} & 83.1 & \textcolor{red}{87.6} & 54.4 & 62.0 & \textcolor{blue}{63.1} & \textcolor{red}{66.6} \\
  & \textcolor{purple}{F} & 77.3 & 79.6 & 77.3 & 74.8 & \textcolor{blue}{87.0} & 81.9 & \textcolor{red}{87.7} & 56.3 & 63.6 & \textcolor{blue}{65.0} & \textcolor{red}{67.3} \\ 
  & $\Delta$Obj & - & - & - & - & - & - & - & 11.7 & 7.7 & \textcolor{red}{4.0} & \textcolor{blue}{4.9} \\ \hline \hline
\multirow{4}{*}{\rotatebox[origin=c]{90}{CB}} & \textcolor{orange}{P} & 84.3 & \textcolor{blue}{87.6} & 79.9 & 86.8 & 74.6 & \textcolor{red}{87.7} & 83.1 & 60.8 & \textcolor{red}{67.6} & \textcolor{blue}{64.9} & 57.7 \\
  & \textcolor{cyan}{R} & \textcolor{blue}{91.7} & 85.0 & 69.3 & 77.5 & 77.0 & \textcolor{red}{93.1} & 89.7 & 44.7 & 58.3 & \textcolor{red}{67.3} & \textcolor{blue}{61.9} \\
  & \textcolor{purple}{F} & \textcolor{blue}{86.6} & 80.6 & 73.7 & 78.2 & 70.5 & \textcolor{red}{90.1} & 83.5 & 45.8 & \textcolor{blue}{60.3} & \textcolor{red}{65.6} & 58.3 \\ 
  & $\Delta$Obj & - & - & - & - & - & - & - & \textcolor{blue}{3.4} & \textcolor{blue}{3.4} & \textcolor{blue}{3.4} & \textcolor{red}{3.2} \\ \hline \hline
\multirow{4}{*}{\rotatebox[origin=c]{90}{CA}} & \textcolor{orange}{P} & \textcolor{red}{81.9} & 73.3 & 82.3 & 77.8 & 77.6 & \textcolor{blue}{80.4} & 78.5 & \textcolor{red}{84.7} & 77.8 & \textcolor{blue}{83.8} & 77.2 \\
  & \textcolor{cyan}{R} & 74.6 & 56.7 & 68.5 & 62.0 & 51.1 & \textcolor{blue}{75.2} & \textcolor{red}{79.7} & 59.4 & 68.1 & \textcolor{blue}{70.0} & \textcolor{red}{77.2} \\
  & \textcolor{purple}{F} & 76.3 & 60.4 & 72.5 & 64.8 & 50.8 & \textcolor{blue}{76.0} & \textcolor{red}{77.1} & 61.5 & 70.0 & \textcolor{blue}{72.2} & \textcolor{red}{75.3} \\ 
    & $\Delta$Obj & - & - & - & - & - & - & - & 22.2 & 5.7 & \textcolor{red}{5.0} & \textcolor{blue}{5.4} \\ \hline \hline
  \multirow{4}{*}{\rotatebox[origin=c]{90}{All}} & \textcolor{orange}{P} & 80.8 & 82.1 & 84.7 & 85.3 & \textcolor{red}{87.4} & 84.7 & \textcolor{blue}{87.1} & 73.8 & \textcolor{blue}{74.5} & \textcolor{red}{75.1} & 74.1 \\
  & \textcolor{cyan}{R} & 80.7 & 75.8 & 73.9 & 70.7 & 77.2 & \textcolor{blue}{82.7} & \textcolor{red}{86.2} & 54.3 & 62.8 & \textcolor{blue}{65.0} & \textcolor{red}{68.2} \\
  & \textcolor{purple}{F} & 78.2 & 75.8 & 75.9 & 73.1 & 77.7 & \textcolor{blue}{81.5} & \textcolor{red}{85.1} & 56.2 & 64.5 & \textcolor{blue}{66.5} & \textcolor{red}{67.9} \\
    & $\Delta$Obj & - & - & - & - & - & - & - & 12.9 & 6.8 & \textcolor{red}{4.1} & \textcolor{blue}{4.8} \\ 
    \hline
\end{tabular}}
\caption{Results for FBMS, ComplexBackground (CB), CamouflagedAnimal (CA), and averaged over all videos in these datasets (ALL). Best results are highlighted in \textcolor{red}{red} with second best in \textcolor{blue}{blue}.}
\label{table:segmentation_results}
\vspace{-3mm}
\end{table*}

\subsection{Ablation Studies} \label{subsec:ablation_studies}

\removed{
\begin{table}
\centering
\small
\begin{tabular}{c||cc||c}
 Dataset & \multicolumn{2}{c||}{DAVIS} & FT3D \\
 Metric & $\mathcal{J}$ & $\mathcal{F}$ & IoU \\ \hline
 NLC \cite{faktor2014video} & 55.1 & 52.3 & - \\
 FST \cite{papazoglou2013fast} & 55.8 & 51.1 & - \\
 CUT \cite{keuper2015motion} & 55.2 & 55.2 & - \\
 FSEG \cite{fusionseg} & 70.7 & 65.3 & - \\
 MPNet \cite{tokmakov2017mpnet} & 70.0 & 65.9 & 85.9 \\
 LVO \cite{tokmakov2017learning} & \textcolor{red}{75.9} & 72.1 & - \\ \hline
 Ours & 74.2 & \textcolor{red}{73.9} & \textcolor{red}{90.7} \\ \hline
\end{tabular}
\caption{Results on Video Foreground Segmentation for DAVIS2016 and FT3D. Best results are highlighted in \textcolor{red}{red}.}
\label{table:video_foreground_segmentation_DAVIS_FT3D}
\vspace{-2mm}
\end{table}
}

\begin{table}[t]
\resizebox{\linewidth}{!}{\begin{tabular}{cc|cccc|c}
 & & FST \cite{papazoglou2013fast} & FSEG \cite{fusionseg} & MPNet \cite{tokmakov2017mpnet} & LVO \cite{tokmakov2017learning} & Ours \\ \hline \hline
\multirow{2}{*}{DAVIS} & $\mathcal{J}$ & 55.8 & 70.7 & 70.0 & \textcolor{red}{75.9} & 74.2 \\
  & $\mathcal{F}$ & 51.1 & 65.3 & 65.9 & 72.1 & \textcolor{red}{73.9} \\ \hline \hline
FT3D & IoU & - & - & 85.9 & - & \textcolor{red}{90.7}
\end{tabular}}
\caption{Results on Video Foreground Segmentation for DAVIS2016 and FT3D. Best results are highlighted in \textcolor{red}{red}.}
\label{table:video_foreground_segmentation_DAVIS_FT3D}
\vspace{-4mm}
\end{table}

\noindent \textbf{Fusion ablation.} We show the choice of mid-level fusion with Y-Net is empirically a better choice than early fusion and late fusion of encoder-decoder networks. For early fusion, we concatenate RGB and optical flow and pass it through a single U-Net. For late fusion, there are two U-Nets: one for RGB and one for optical flow, with a conv layer at the end to fuse the outputs. Note that Y-Net has more parameters than early fusion but less parameters than late fusion. \removed{We train all three models with $C=32$ on FT3D, DAVIS, and FBMS. }Table \ref{table:fusion_ablation} shows that Y-Net outperforms the others in terms of foreground IoU. Note that the performance gap is more prominent on the real-world datasets. \removed{All models were trained on FT3D for 100k iterations, and fine-tuned on the real-world dataset for 50k iterations.}

\vspace{2mm}
\noindent \textbf{Architecture ablation.} We evaluate the contribution of each part of the model and show results in both the multi-object setting and the binary setting (foreground segmentation) on the FBMS testset. All models are pre-trained on FT3D for 150k iterations and trained on FBMS+DAVIS-m for 100k iterations. Experiments with the different PT-RNN variants shows that {\it conv} PT-RNN performs the best empirically in terms of F-score, thus we use this in our comparison with state-of-the-art methods. {\it Standard} performs similarly, while {\it convGRU} performs worse perhaps due to overfitting to the small dataset. Next, we remove the PT-RNN architecture (per-frame embedding) and cluster the foreground pixels at each frame. \removed{Cluster matching is done with a window size of 1. }The F-score drops significantly and $\Delta$Obj is much worse, which is likely due to this version not labeling clusters consistently in time. Because the foreground prediction is not affected, these numbers are still reasonable. Next, we remove foreground masks (no FG mask) and cluster all foreground and background trajectories. The clustering is more sensitive; if the background trajectories are not clustered adequately in the embedding space, the performance will suffer. Lastly, we removed the spatial coordinate module (no SCM) and observed lower performance. Similar to the per-frame embedding experiment, foreground prediction is not affected.

\vspace{2mm}
\noindent \textbf{Dataset ablation.} We also study the effects of the training schedule and training dataset choices. In particular, we first explore the effect of not pre-training on FT3D, shown in the bottom portion of Table \ref{table:architecture_and_dataset_ablation}. Secondly, we explore the effect of training the model only on FBMS (without DAVIS-m). Both experiments show a noticeable drop in performance in both the multi-object and foreground/background settings, showing that these ideas are crucial to our performance. 

\begin{figure*}[t]
\begin{center}
\includegraphics[width=\linewidth]{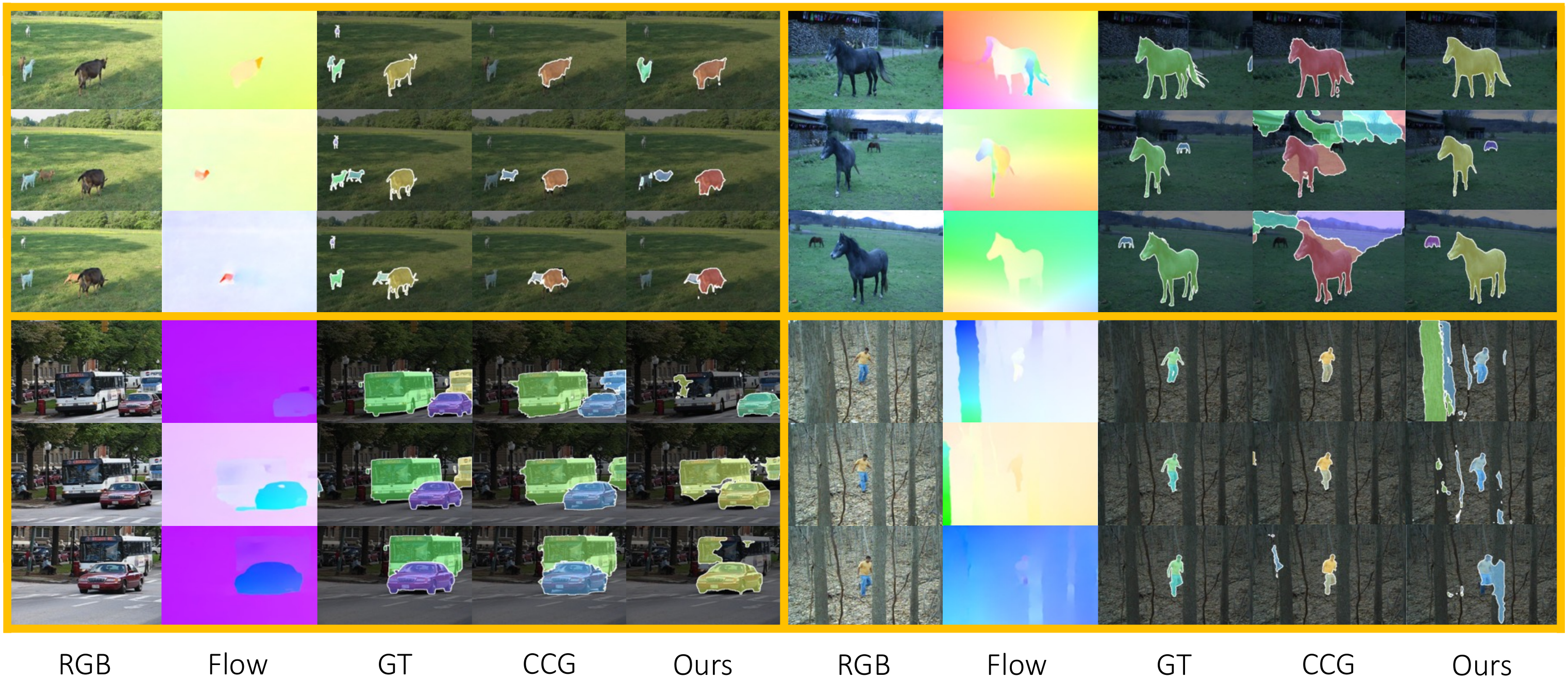}
\caption{Qualitative results for motion segmentation. The videos are: {\it goats01}, {\it horses02}, and {\it cars10} from FBMS, and {\it forest} from ComplexBackground.}
\label{fig:qualitative_motion_segmentation}
\vspace{-8mm}
\end{center}
\end{figure*}

\subsection{Comparison to State-of-the-Art Methods}
\label{subsec:SOTA_comparison}

% We show results of our method against state-of-the-art methods in video foreground segmentation and multi-object motion segmentation. 

%As discussed in the previous section, we select our full model to be the {\it conv} PT-RNN variant of Figure \ref{fig:pt_rnn_architecture}. We train this model for 150k iterations on FT3D, then fine-tune on FBMS+DAVIS-m for 100k iterations. 

\vspace{2mm}
\noindent \textbf{Video Foreground Segmentation.} For FBMS, ComplexBackground and CamouflagedAnimal, we follow the protocol in \cite{bideau2018best} which converts the motion segmentation labels into a single foreground mask and use the metrics defined in \cite{ochs2014segmentation} and report results averaged over those three datasets. We compare our method to state-of-the-art methods including PCM \cite{bideau2016s}, FST \cite{papazoglou2013fast}, NLC \cite{faktor2014video}, MPNet \cite{tokmakov2017mpnet}, LVO \cite{tokmakov2017learning}, and CCG \cite{bideau2018best}. We report results in Table \ref{table:segmentation_results}. In terms of F-score, our model outperforms all other models on FBMS and CamouflagedAnimal, but falls just short on ComplexBackground behind PCM and CCG. Looking at all videos, we show a relative gain of 4.4\% on F-score compared to the second best method CCG, due to our high recall.

Additionally, we report results of our model on FT3D and the validation set of DAVIS2016. We compare our model to state-of-the-art methods: including LVO \cite{tokmakov2017learning}, FSEG \cite{fusionseg}, MPNet \cite{tokmakov2017mpnet}, and FST \cite{papazoglou2013fast} in Table \ref{table:video_foreground_segmentation_DAVIS_FT3D}. For this experiment only, we train a Y-Net with $C=64$ channels on FT3D for 100k iterations, resulting in outperforming MPNet by a relative gain of 5.6\%. We then fine-tune for 50k iterations on the training set of DAVIS2016 and use a CRF \cite{krahenbuhl2011efficient} post-processing step. We outperform all methods in terms of $\mathcal{F}$-measure and all methods but LVO on $\mathcal{J}$-measure. Note that unlike LVO, we do not utilize an RNN for video foreground segmentation, yet we still achieve performance comparable to the state-of-the-art. Also, LVO \cite{tokmakov2017learning} reports a $\mathcal{J}$-measure of 70.1 without using a CRF, while our method attains a $\mathcal{J}$-measure of 71.4 without using a CRF. This demonstrates the efficacy of the Y-Net architecture.

\vspace{1mm}
\noindent \textbf{Multi-object Motion Segmentation.} We compare our method with state-of-the-art methods CCG \cite{bideau2018best}, CUT \cite{keuper2015motion}, and CVOS \cite{taylor2015causal}. We report our results in Table \ref{table:segmentation_results}. We outperform all models on F-score on the FBMS and CamouflagedAnimal datasets. On FBMS, we dominate on precision, recall, and F-score with a relative gain of 3.5\% on F-score compared to the second best method CCG. Our performance on $\Delta$Obj is comparable to the other methods. On CamouflagedAnimal, we show higher recall with lower precision, leading to a 4.4\% relative gain in F-score. Again, our result on $\Delta$Obj is comparable. However, our method places third on the ComplexBackground dataset. This small 5-sequence dataset exhibits backgrounds with varying depths, which is hard for our network to correctly segment. However, we still outperform all other methods on F-score when looking at all videos. Similarly to the binary case, this is due to our high recall. Because we are the first work to use FT3D for motion segmentation, we report results on FT3D in the supplement for the interested readers.

To illustrate our method, we show qualitative results in Figure \ref{fig:qualitative_motion_segmentation}. We plot RGB, optical flow \cite{IMKDB17}, groundtruth, results from the state-of-the-art CCG \cite{bideau2018best}, and our results on 4 sequences ({\it goats01}, {\it horses02}, and {\it cars10} from FBMS, and {\it forest} from ComplexBackground). On {\it goats01}, our results illustrate that due to our predicted foreground mask, our method is able to correctly segment objects that do not have instantaneous flow. CCG struggles in this setting. On {\it horses02}, we show a similar story, while CCG struggles to estimate rigid motions for the objects. Note that our method provides accurate segmentations without the use of a CRF post-processing step. We show two failure modes for our algorithm: 1) if the foreground mask is poor, the performance suffers as shown on {\it cars10} and {\it forest}, and 2) cluster collapse can cause multiple objects to be segmented as a single object as shown in {\it cars10}.

\vspace{-2mm}
\section{Conclusion}
\vspace{-2mm}
We proposed a novel deep network architecture for solving the problem of object discovery using object motion cues. We introduced an encoder-decoder network that learns representations of video frames and optical flow, and a novel recurrent neural network that learns feature embeddings of pixel trajectories inside foreground masks. By clustering these embeddings, we are able to discover and segment potentially unseen objects in videos. We demonstrated the efficacy of our approach on several motion segmentation datasets for object discovery.

\vspace{-2mm}
\section*{Acknowledgements}
\vspace{-1mm}

We thank Pia Bideau for providing evaluation code. This work was funded in part by an NDSEG fellowship, the program ``Learning in Machines and Brains'' of CIFAR, and faculty research awards.

{\small
\bibliographystyle{ieee}
\bibliography{object_discovery}
}

%\newpage
%\input{appendix}

\end{document}